# A Comparative Evaluation of Prominent Methods in Autonomous Vehicle Certification[1]

Mustafa Erdem Kırmızıgül[1,2,3*], Hasan Feyzi Doğruyol[4], Haluk Bayram[1,2]

[1] Field Robotics Laboratory, Science and Advanced Technologies Research Center (BILTAM), Istanbul Medeniyet University, Istanbul, Turkey

[2] Field Robotics Laboratory, Robotics and Artificial Intelligence Laboratories (ROYAL), Bogazici University, Istanbul, Turkey

[3] Directorate General for Regulation of Transport Services, Ministry of Transportation and Infrastructure, Ankara, Turkey

[4] HAVELSAN, Technology Development Zone, 41400, Gebze, Kocaeli, Turkey

**Abstract**

The "Vision Zero" policy, introduced by the Swedish Parliament in 1997, aims to eliminate fatalities and serious injuries resulting from traffic accidents. To achieve this goal, the use of self-driving vehicles in traffic is envisioned and a roadmap for the certification of self-driving vehicles is aimed to be determined. However, it is still unclear how the basic safety requirements that autonomous vehicles must meet will be verified and certified, and which methods will be used. This paper focuses on the comparative evaluation of the prominent methods planned to be used in the certification process of autonomous vehicles. It examines the prominent methods used in the certification process, develops a pipeline for the certification process of autonomous vehicles, and determines the stages, actors, and areas where the addressed methods can be applied. Figure A illustrates a general view of a certification process.

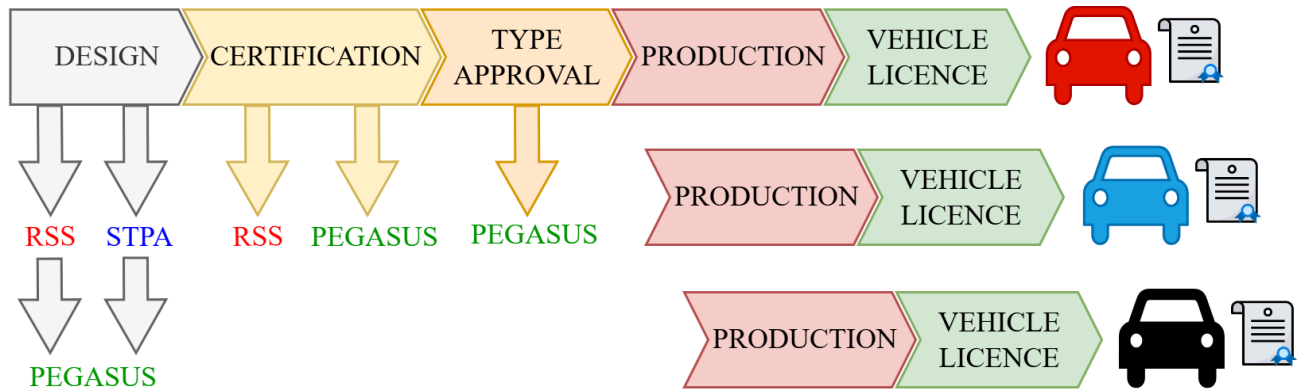

**Figure A.** Pipeline in the Certification Process of Autonomous Vehicles

**Highlights**

- Comparative analysis of RSS, PEGASUS and STPA Methods.
- Evaluation of these methods and their roles in the certification process of autonomous vehicles.
- Incorporating these methods into the certification process and developing a flow-chart.

---

**Summary**

The "Vision Zero" policy, introduced by the Swedish Parliament in1997, aims to reduce traffic fatalities and serious injuries to zero. To achieve this goal, the policy envisions the use of self-driving vehicles in traffic and seeks to establish a roadmap for their certification. However, there are still unclear points, such as: What basic safety requirements must an autonomous vehicle meet to be certified? How can these requirements be verified and certified, and which methods can be used? This paper examines the comparative evaluation of certification methods (Responsibility-Sensitive Safety - RSS, System-Theoretic Process Analysis - STPA and PEGASUS) for autonomous vehicles. Figure A illustrates a general view of a certification process.

**Purpose:** The paper aims to provide a comparative evaluation of the certification methods for autonomous vehicles and to determine which method should be used at specific stages, by which actors, and for what purposes within the certification process.

**Theory and Methods:** Our study adopts a comprehensive and comparative approach to evaluate the certification methods for autonomous vehicles. Grounded in the critical aspects of safety assessment, validation, and certification, we examine and compare prominent methods, including RSS, STPA, and PEGASUS. Through a rigorous evaluation, we assess the strengths and limitations of each method, considering their applicability at different stages, by diverse actors, and for specific purposes within the certification process. Furthermore, we develop a structured pipeline model that depicts the sequential steps involved in the certification process of autonomous vehicles.

**Results:** After conducting the study, we evaluate the certification process for autonomous vehicles and analyze the roles of various methods in a comparative manner. Each method is positioned within a pipeline representing the autonomous vehicle certification process.

**Conclusion:** In this paper, we address the critical aspects of safety assessment, validation, and certification for autonomous vehicles. Specifically, we conduct a comparative examination of prominent methods such as RSS, STPA, and PEGASUS in the context of autonomous vehicle certification. Furthermore, we establish a structured pipeline for the certification process of autonomous vehicles, as depicted in Figure A. Additionally, we identify the appropriate method to be employed at different stages, by various actors, and for specific purposes within the process.

# 1. INTRODUCTION

In general, an autonomous vehicle is defined as a vehicle that has the ability to sense its environment and travel without human intervention [1]. In the J3016 standard, the third revision of which is currently being published by the Society of Automotive Engineers (SAE), a taxonomy is defined in which vehicles are divided into six autonomy levels according to their autonomy capabilities. With this taxonomy, vehicles are divided into non-autonomous driving, driver assistance, partially autonomous driving, conditional autonomous driving, highly autonomous driving and fully autonomous driving levels from level 0 to level 5 [2].

According to data published by the National Highway Traffic Safety Administration (NHTSA), 42,939 people lost their lives in road accidents in the United States in 2021 [3]. Considering this remarkable data, the use of fully autonomous vehicles independent of human control becomes very important in order to reduce accidents caused by human error to zero. Today, considering that some countries have already completed studies such as determining the necessary test scenarios, establishing test centers [4] and preparing verification protocols [5] for the launch of autonomous vehicles, and that vehicles at the 3rd and 4th autonomy levels have been put into use, it is clear that we are about to see 5th level autonomous vehicles in traffic. One of the important problems to be overcome in order for autonomous vehicles, which are considered to have completed their technical development at a sufficient level, to be put into use is the harmonization of the road infrastructure [6]. Another problem is to make new traffic regulations that will guarantee traffic safety by taking into account the traffic behavior of autonomous vehicles [7, 8]. Another problem is to determine the minimum technical and functional requirements that the vehicle should provide, and to clarify the process and methods of conformity assessments for testing, certification and type approval.

Schöner, who defined this problem in the most basic form in 2016, stated that test scenarios and procedures are needed for autonomous vehicles to be accepted as safe [9]. Junietz et al. emphasized that the vehicle industry as well as the authorities that will authorize the autonomous vehicle to be put into service are stakeholders in the testing of the autonomous vehicle, and the test method to be adopted must be acceptable by both the industry and the authorities and proposed a scenario-based testing approach [10]. In order to prepare the safety standard for autonomous vehicles, Koopman et al. examined different standards that are currently valid for different subjects, such as the safety of conventional vehicles or railway signaling systems, but basically developed to guarantee safety, and drew the framework of the safety standard for autonomous vehicles [11]. On the other hand, the European Commission report prepared by Baldini formally addressed the concept of testing and certification of autonomous vehicles and recommended that in order to test the safety of the vehicle, the safety of the software that will manage the autonomy functions should also be tested, and cyber security issues should also be clarified [12].

On the legal side, limited regulations have been made regarding the procedures to be taken as basis for type approval assessments in order to enable the use of vehicles' autonomous functions within predefined areas and routes, subject to restrictions, before the European Commission and the United Nations Economic Commission for Europe (UNECE) [13, 14, 15]. Nevertheless, the regulations adopted to date do not fulfill the requirements for an internationally accepted certification process that would allow autonomous vehicles to be used without restriction in international traffic. For this reason, the stakeholders in the road vehicle and infrastructure sector, especially the public authorities, are following the work carried out by the international regulatory authorities and are expected to develop and accept test methodologies that will respond to the regulatory needs at the international level. In this process, Sweden has decided to determine a road map for the use of autonomous vehicles in order to reduce accidents caused by human error in traffic to zero with "Vision Zero" policy [16]. France continues to prepare local legislation [17]. Germany is encouraging scientists and companies producing automobiles and automotive materials to come together and contribute to addressing this need [18].

To date, many studies have been conducted to develop methods for determining and testing the safety requirements of autonomous vehicles. However, there has not been a comprehensive study explaining which of the currently developed methods will be used by which of the actors taking part in the certification process and for what purpose, from the design to type approval of the autonomous vehicle. On the other hand, with the RSS, STPA and PEGASUS methods described in the "New Approaches for the Certification of Autonomous Vehicles" report published by the European Commission in 2020 [19], prominent methods have been brought together to respond to the needs of all stakeholders playing a role in the certification and type approval process. In this article, we examine these methods in detail and propose a taxonomy - a hierarchical model of the entire process from vehicle design to type approval. This taxonomy specifies which methods will be utilized by each actor involved and for what purpose.

## II. THE PROMINENT METHODS FOR CERTIFICATION OF AUTONOMOUS VEHICLES

In this section, the RSS, PEGASUS and STPA methods addressed in the European Union's "New Approaches for the Certification of Autonomous Vehicles" report for definition of the safety requirements of autonomous vehicles, verification and certification according to these requirements are examined and evaluated [19].

**2.1 Responsibility Sensitive Safety (RSS)**

The RSS method was first developed by Mobileye in 2017 to respond to two fundamental problems in order to guarantee the safety of autonomous vehicles [20]. The first one is what are the basic safety requirements that an autonomous vehicle must meet. The second is how to verify that the autonomous vehicle meets the basic safety requirements. The remarkable promise of the developers of RSS is to propose clear mathematical models to respond to these questions. The mathematical models identified through RSS are based on Newtonian mechanics [21]. These mathematical models will allow not only the verification of the basic safety requirements of the autonomous vehicle, but also the grading of its ability to meet these requirements. These mathematical models described by the RSS method will be used as a basis for determining the linear and lateral safety distances, which are the basic elements of traffic safety.

In traffic, there are written and unwritten rules that every driver must follow. Technically, these rules can be classified as explicit rules and implicit rules [22]. Explicit rules are the rules set by official authorities through legal regulations. Speed limits or traffic signs can be given as examples to these kinds of rules. Implicit rules are not determined by any written source, may vary culturally, and have emerged over time according to general acceptance, shaped according to the habits and driving behaviors of vehicle users in traffic. Although implicit rules are not as precise and strictly enforced as legal regulations, they play an important role in ensuring the driving safety of vehicles in traffic. It may be considered easier to design an autonomous vehicle, which is essentially a robot, to comply with explicit rules and to test and guarantee that it complies with these rules. However, testing according to unwritten implicit rules will bring many more problems to be overcome. To overcome these problems, the implicit rules need to be generalized and made explicit.

On the other hand, the RSS method has been criticized for formulating only lateral and linear safety distances into mathematical models is not enough to guarantee safety. Accordingly, an autonomous vehicle, which is a robot, can only obey the rules based on its design. However, a human driver can take the initiative to disobey the rules, predicting that following the rules may result in danger in some situations [23]. For example, when a human driver notices another vehicle that is potentially in collision or an object that has fallen on the road, the driver may accelerate to avoid a collision, risking exceeding the speed limit, or may react by changing lanes more than once without risking safety.

In order to respond to all these different approaches, Mobileye has not only formulated lateral and linear safety distances with the Responsibility Sensitive Safety (RSS) method, but also generalized the implicit rules and proposed five rules that the autonomous vehicle must comply with and explained the metrics by which the autonomous vehicle's ability to comply with these rules will be tested [24].

In this paper, the usage of the method in the design and testing of autonomous vehicles is addressed, but RSS has also been used in the design of driving assistance systems for conventional vehicles. For example, Chai et al. used the RSS method in the development of an adaptive cruise control system [24]. Koopman et al. and Liu et al. clearly explain how the method can be used to calculate lateral and linear safety distances for the design of autonomy functions of autonomous vehicles [21, 25].

**2.1.1 Five safety rules of autonomous vehicles**

In the RSS method, five scenario-based rules are developed to guarantee the safety of an autonomous vehicle. These rules will be explained in this section.

**Rule 1: Do not hit the vehicle in front**

The first principle of the RSS method is to maintain a safe following distance between two following vehicles in order to guarantee that the two vehicles do not collide with each other. The developers of the RSS method define the safe following distance as the minimum distance at which the following vehicle can stop without colliding in the event of a sudden braking stop of the lead vehicle [22]. Figure 1 shows the safe following distance determined by the sum of the reaction time and the stopping time.

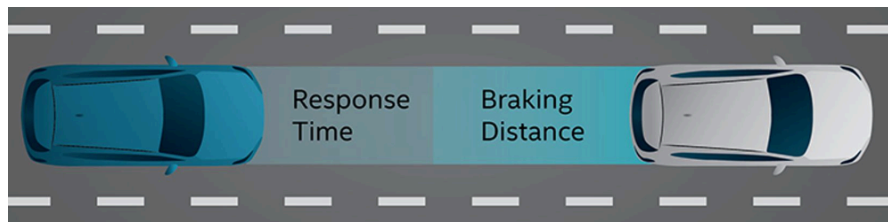

**Figure 1.** Response Time and Braking Distance [22].

It is generally accepted by vehicle users that the safe following distance between two vehicles is determined by the two-second rule. Accordingly, when the driver of the following vehicle determines the safe following distance from the vehicle in front of him, he adjusts the speed of his vehicle so that he reaches a point where the vehicle in front of him passes after 2-3 seconds. In Figure 2, the safe following distance is represented by $d_{min}$, which is the sum of the reaction time and braking distance. In this practical RSS method, it is formulated as in Equation 1 [22].

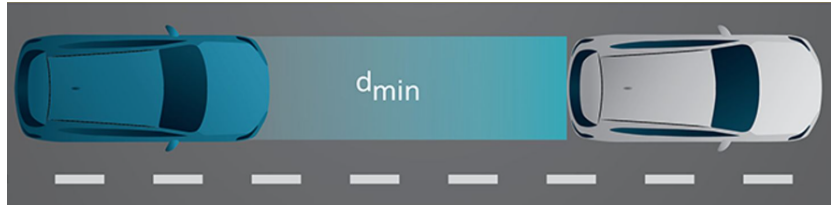

**Figure 2.** Safe Longitudinal Distance [22].

$$d_{min} = [V_r P + \frac{1}{2}\alpha_{max} P^2 + \frac{(V_r + P\alpha_{max})^2}{2\beta_{min}} - \frac{V_f^2}{2\beta_{maks}}]_+ \quad (1)$$

- $\alpha_{max}$ maximum permissible acceleration for the vehicle behind within the reaction time
- $\beta_{max}$ maximum allowable deceleration for the vehicle in front
- $\beta_{min}$ minimum allowable deceleration for the vehicle in front
- $d_{min}$ minimum safe following distance
- $P$ reaction time of the vehicle behind
- $V_r$ speed of the rear vehicle
- $V_f$ speed of the front vehicle

According to Equation 2.1, as soon as the distance between the autonomous vehicle and the vehicle in front of it is less than the distance calculated by $d_{min}$, the vehicle will react accordingly and slow down, and the distance between the vehicle and the vehicle in front of it will be increased back above $d_{min}$.

**Rule 2: Do not cut in recklessly**

The vehicle must not violate the defined safe lateral distances both when traveling in its lane and when changing lanes. For example, when traveling in a lane, it must not approach the lane lines and must center the lane as much as possible. In the case of overtaking, they must maintain safe lateral distances when approaching or overtaking the overtaken vehicle and must not get closer. Likewise, if a vehicle is overtaken, the overtaken vehicle should approach to the right within its lane in order to prevent a possible collision and reduce its speed by braking to safely move away from the overtaking vehicle. Thus, it is necessary to determine how to calculate safe lateral distances. The developers of the RSS method explain that the safe lateral distance shown in Figure 3 are calculated as in Equations 2 and 3 [22].

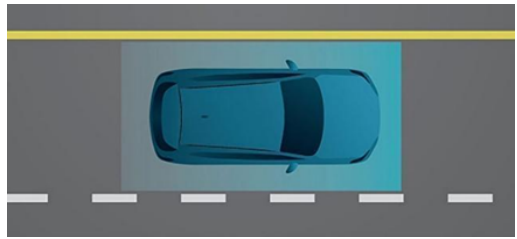

**Figure 3.** Safe Lateral Distance [22].

$$v_{1,p} = v_1 + P\alpha_{lateral,max,acceleration}, \quad v_{2,p} = v_2 - P\alpha_{lateral,max,acceleration} \quad (2)$$

$$d_{lateral,min} = \mu + [\frac{(V_1 + V_{1,P})}{2}P + \frac{V_{1,P}^2}{2\beta_{1,lateral,min}} - ((\frac{V_2 + V_{1,P}}{2})P + \frac{V_{2,P}^2}{2\beta_{2,lateral,min}})]_+ \quad (3)$$

- α $_{lateral,max,acceleration}$ maximum lateral acceleration of the vehicle
- $β_{1,lateral,min}$ ve $β_{2,lateral,min}$ minimum deceleration of the vehicle to the right and left
- $μ$ safe minimum lateral distance
- $d_{lateral,min}$ minimum safe lateral distance
- $V_1$ ve $V_2$ left and right lateral velocities of the vehicle
- $V_1,p$ and $V_2,p$ lateral velocities of the vehicle within the response time
- $P$ response time of the vehicle

**Rule 3: Right of way is given, not taken**

For roads with clear road markings and complete traffic signs and markers, the priorities of passage at intersections and junctions are clearly defined. Drivers know which vehicle has the right of way in line with the directions of these signs and markers. However, not all roads have road markings, traffic signs and markers to clearly identify which vehicle has the right of way. For example, in Figure 4, there should be a stop sign for the driver exiting the secondary road onto the main road, but there is no such sign for the drivers of both vehicles.

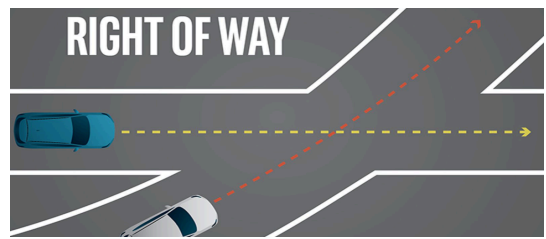

**Figure 4.** Right of Way [22].

If both roads have equal status, the right of way would belong to the vehicle on the right. However, if the road used by the vehicle on the left is the main road, then the right of way should belong to the vehicle on the left. In this case, it is not clear which vehicle has the right of way. In practice, the right of way on such roads is determined by agreement between the drivers of the two vehicles approaching the intersection. However, an autonomous vehicle will not be able to make such an agreement. In this case, a method should be used to determine which vehicle has the right of way. On the other hand, even if the right of way belongs to the autonomous vehicle, the other vehicle may violate traffic rules and take the road. In this case, even if the right of way belongs to the autonomous vehicle, it is not acceptable for the autonomous vehicle to continue on the road and cause an accident. Considering these kinds of situations, according to the RSS method, it is accepted that the principle of "right of way is given, not taken" will be valid for the safety of traffic.

**Rule 4: Be cautious in limited visibility conditions**

Limited visibility can be caused by many things, from the topology of the road to buildings on the side of the road, to adverse weather conditions, to other vehicles in traffic. In such conditions, vehicle drivers may not be able to recognize when there is a risky situation ahead. The same is true for autonomous vehicles. Even though it is against traffic rules, it is a very serious danger for pedestrians to attempt to cross the road without a pedestrian lane. On major intercity roads, for example, this may seem improbable. But in residential neighborhoods, this possibility must always be considered. It is also possible that a part of the pedestrian lane is not visible, as shown in Figure 5.

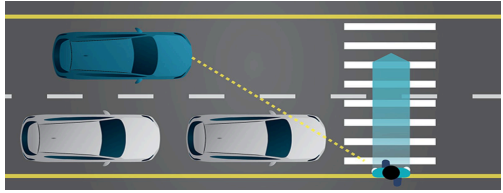

**Figure 5.** Areas of Limited Visibility [22].

In particular, it is possible for vehicles to be parked on the side of the road so as to block the view of a pedestrian, or even if other vehicles are moving, they may block the view of a pedestrian. In such cases, drivers should be very cautious, especially when approaching pedestrian lanes, as they cannot see pedestrians who will try to cross the road in advance. Autonomous vehicles should act with the same assumption.

**Rule 5: If you can avoid an accident without causing another accident, do so**

The first four rules explain the scenarios that may cause a dangerous situation and how to take precautions for these situations. The fifth rule explains when an accident is unavoidable due to a dangerous situation that occurs unexpectedly and suddenly. According to the fifth rule, if an accident is unavoidable, if the autonomous vehicle can avoid it without causing another accident, it should do so. Figure 6 illustrates such a situation. If the blue autonomous vehicle tries to move into the right lane to avoid hitting objects falling from the vehicle in front, it will cause another accident with the vehicle just to its right. According to the RSS method, this situation is not acceptable, and the vehicle should, if possible, move to a safe area without causing an accident and, if not possible, apply full break.

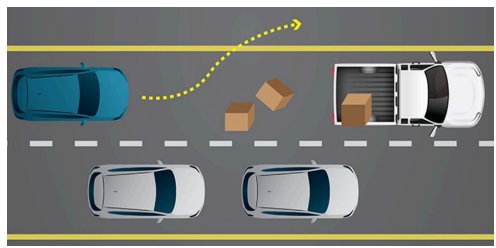

**Figure 6.** Avoid Collision Without Causing Another Collision [22].

**2.1.2 Implementation of RSS Method**

The research of National Highway Traffic Safety Administration (NHTSA) has identified 37 different pre-crash scenarios that are involved in 99.4% of light vehicle crashes [26]. Some of these scenarios are related to driver behavior, while others are not caused by driver fault, such as vehicle malfunctions or tire blowouts. Some of the driver-fault accidents are accidents in which only the driver of the vehicle involved in the accident is at fault and no other vehicle plays a role, while others are accidents in which more than one vehicle driver plays a role. The RSS only addresses accident scenarios in which more than one driver plays a role. For this reason, the implementation of the RSS method is described for pre-crash scenarios where more than one driver is involved, among the scenarios identified by NHTSA. These scenarios are described below.

**One way traffic scenarios**

One-way scenarios based on the movement of the vehicle ahead are shown in Figures 7 and 8.

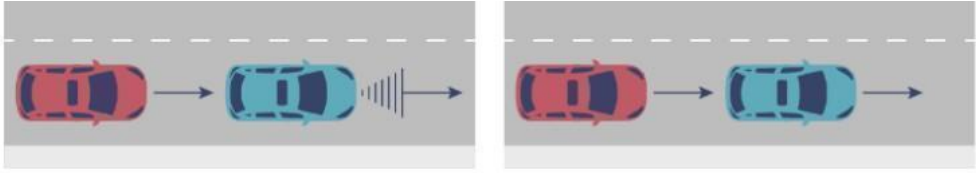

**Figure 8.** Acceleration of the Front Vehicle (Left), Slower Movement of the Front Vehicle (Right) [26].

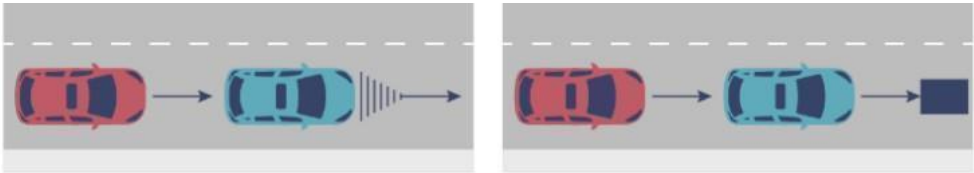

**Figure 8.** Deceleration of the Leading Vehicle (Left), Stopping of the Leading Vehicle (Right) [26].

Using the RSS method, the four different one-way scenarios above are quite similar to each other, so all four scenarios are considered as a single scenario.

**Lane change and drift scenarios**

Scenarios based on lane changing and drifting are shown in Figures 9 and 10.

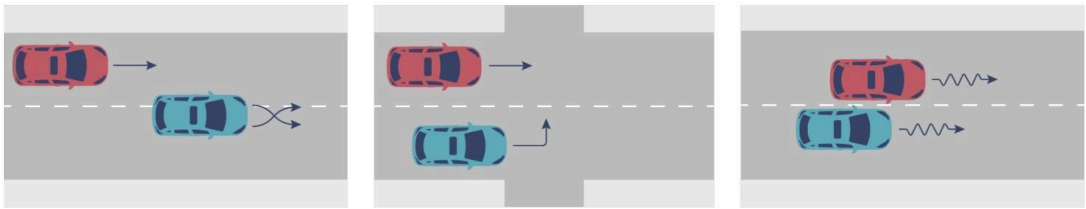

**Figure 9.** Lane Changing (Left), Wrong Lane Turning (Center), Drifting (Right) [26].

In the first and second scenarios in Figure 9, it is necessary to evaluate whether the lane change is performed safely or not according to the scenarios in Figure 10.

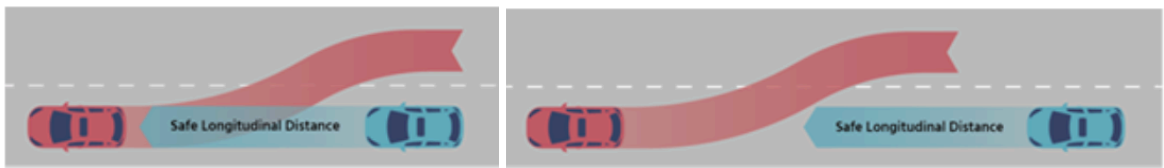

**Figure 10.** Unsafe Lane Change (Left), Safe Lane Change (Right) [26].

In Figure 10, in the first case, the red vehicle has switched lanes and moved into the lane of the blue vehicle, overlapping the safe stopping distance of the blue vehicle behind. In such a scenario, if the red vehicle makes a sudden stop within the safe stopping distance of the blue vehicle, it may cause an accident. Therefore, this lane change is an unsafe lane change. In the second case, since the red vehicle enters the blue vehicle's lane after passing the blue vehicle's safe stopping distance, there is no risk of an accident in case of a sudden stop. This lane change is a safe lane change.

**2.2 PEGASUS Method**

Initial work on the PEGASUS method started in January 2016 and was supported by the German Federal Government [18,27]. The ongoing project aims to identify the basic safety criteria that autonomous vehicles must meet, the tools, test methods and scenarios that will be required to test these criteria, with the participation of leading companies in the German automobile industry and academia. The main challenge to be overcome here is the need to use different test scenarios for different regions, as the basic safety requirements may vary regionally. The International Alliance for Mobility Testing and Standardization (IAMTS) points out that conditions such as population densities, road infrastructure network, traffic density and traffic rules that differ between regions will bring great differences to the test scenarios to be used in testing autonomous vehicles [28]. In the PEGASUS project, this issue was taken into account and a mechanism was defined that enables the development of different scenarios in line with regional conditions and combines the type approval processes for testing and market launch with a dynamic structure.

PEGASUS is a project initiated in 2016 under the direction of the German Federal Ministry of Economics and Energy (Bundesministerium für Wirtschaft und Energie - BMWi) and carried out with the participation of leading companies in the automotive industry and actors working in the fields of autonomous vehicle technology [27]. This project describes the processes of design, realization, testing, verification and type approval by the authorities for the safety of vehicles at levels 3 and 4 according to the autonomous driving levels determined by SAE.

The implementation of the project is dependent on the enforcement of regulations that will lay down the procedures and principles for type approval requirements and processes. Currently, (EU) 2022/1426 and (EU) 2022/2236 regulations of the European Union (EU) have regulated which subsystems shall be evaluated for the type approval of autonomous vehicles and the evaluation criteria [13,14]. However, the scope of these regulations allows the use of autonomous vehicles in predetermined areas, limited to certain routes, and autonomous parking in designated parking areas. At this stage, in order for autonomous vehicles to be used in international traffic, regulatory studies on internationally valid traffic safety, traffic rules and certification of autonomous vehicles are still ongoing within the WP.29 (World Forum for Harmonization of Vehicle Regulations) and WP.1 (Global Forum for Road Traffic Safety) regulatory forums within the United Nations Economic Commission for Europe (UNECE) [29]. On the other hand, with the ISO 34502:2023 standard, the concept of operational design space for determining the scenarios required to test the compliance of the vehicle with the ongoing regulations and testing the vehicle according to these scenarios was explained, thus completing the first phase of standardization studies [30]. With the finalization of the ongoing standardization and regulation studies, it is considered that the outputs of the PEGASUS project will be put into practice on the basis of legal regulations.

The main output of the PEGASUS project is to establish the standards and verification methods necessary to guarantee safety in autonomous use mode. The aim of the project is represented by the meaning of the letters that make up the word PEGASUS, which also gives the project its name; "project for the establishment of generally accepted quality criteria, tools and methods as well as scenarios and situations for the release of highly automated driving functions".

The main objectives of the project can be explained as follows [18]:
- Defining a standardized procedure for testing autonomous vehicle systems in simulation, test stand and real traffic conditions,
- Development of continuous and variable vehicles to maintain autonomous driving,
- Integration of testing in the initial development process,
- Develop a methodology among manufacturers to maintain the safety of highly autonomous driving functions.

Today, because of the studies on autonomous vehicle technology, the technical requirements needed to fulfill autonomous functions have been met and autonomous cars for private use, autonomous vans for public transportation and even construction machines can be put into service [31]. However, there are still many questions that need to be answered before autonomous vehicles can be used in traffic. Within the scope of the PEGASUS project, the issues that need to be answered regarding the safety of the autonomous vehicle are addressed with the following two basic questions: 1) What are the safety requirements for autonomous vehicles? 2) How can it be proven that these requirements are met?

As is well known, the autonomy function requires the autonomous vehicle to be able to manage itself without any supervision or intervention from the driver. But now that there is no longer any human decision behind the wheel and all responsibility is transferred to the vehicle, the vital importance of this responsibility places heavy requirements that must be met. So, what will be the role of humans in autonomous driving in the future? How can the division of tasks between vehicles and human beings be optimized? Considering that there is an expectation that autonomous vehicle technology will be ready and available as quickly as possible, these issues, which still require intensive research and development work, need to be completed quickly.

New standards and methods need to be developed in close cooperation between the research and industry sectors to confirm the safety of autonomy functions before autonomous vehicles can be deployed. In the Global Road Safety Manifesto published by the International Organization of Motor Vehicle Manufacturers (OICA), it is indicated that international or regional safety requirements for road vehicles should be determined within the scope of international regulations to be prepared through the work carried out under UNECE WP.29 [32]. The PEGASUS joint project will be able to provide solutions to international regulatory studies at this point and will determine the general safety requirements for fully autonomous driving functions and the tools and scenarios to be used to test these requirements. On the other hand, there is a need for a method to validate these test subjects that is acceptable to authorities, manufacturers and the public.

One of the two most critical features of the PEGASUS method is that it is a method that includes all the design, production and final control processes from the production of the vehicle until it is put into service. This process consists of the verification of the design by the manufacturer through analysis and simulations, the production of prototypes, and then the verification of the safety of the designed and produced vehicle through scenario-based tests. In this respect, the PEGASUS method ensures that the roles of both the manufacturer and the safety assessment bodies and the requirements of all stages are clearly defined.

The second critical feature of the PEGASUS method is the consideration of the driving abilities and preferences of the drivers in traffic. The PEGASUS method does not predefine threshold driving skill requirements to be exceeded at the beginning of the assessment process; instead, as shown in Figure 11, drivers' performance and behavior in traffic are collected in a data repository and statistically evaluated to determine threshold driving skill requirements for success or failure based on statistical data [33].

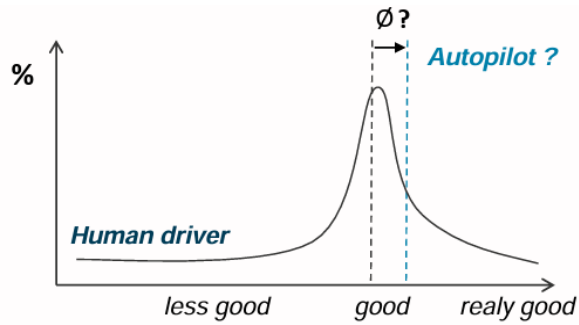

**Figure 11.** Driving Skills [33].

The regional variation in driver performance and behavior means that evaluation criteria should be determined according to the countries where the vehicle will be used, and that separate test scenarios can be used for each country in accordance with the practices of the field of use in order to make a more realistic assessment of the autonomous vehicle. Although the theoretical development of this method has been completed, work is still ongoing on reflecting driver performance and behavior in data and collecting them in data pools.

It is not proper to consider the PEGASUS method as a method developed solely for measuring the safety level of an autonomous vehicle. This method describes a process that starts with the design of the autonomous vehicle, prototype production, testing and analysis, and production, and extends from the design of the autonomous vehicle to the user. In this process, it is a method that can include the roles of the vehicle's designer, subsystem providers, main integrator and manufacturer, test centers, conformity assessment bodies that will control this whole process, official authorities and finally users. In this respect, the PEGASUS method not only defines the safety requirements and measurement parameters, but also explains their testing phases, clarifying the roles and responsibilities of all stakeholders involved. BMWi is developing the PEGASUS method for this very purpose by involving many companies involved in vehicle production in this process.

To explain the PEGASUS method, the process has been transformed into a detailed flowchart as shown in Figure 12. Here, the path of the process is described as data collection and requirements definition, database preparation, assessment of autonomous functions and conclusion.

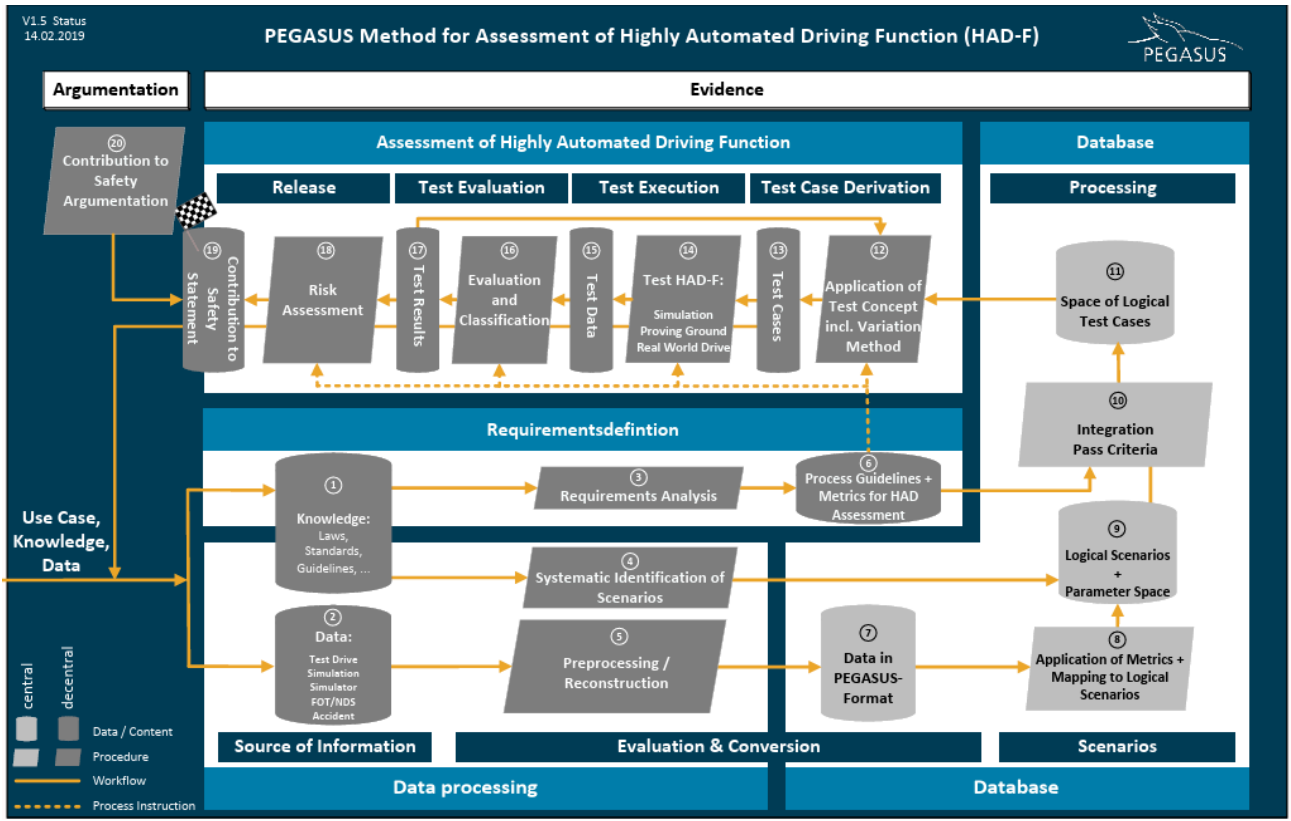

**Figure 12.** Flow Chart of the PEGASUS Method [33].

The steps shown in Figure 12 in Form's work published in 2018, explaining the flow chart of the PEGASUS method are described as follows [27].

**2.2.1 Data processing**

The process represented by Steps 1, 2, 4 and 5 is the data processing process. In the data processing process, the legal requirements that the autonomous vehicle must meet, the technical specifications of the vehicle and the measurement data to be taken as reference for testing its functions and the scenarios to be used in the tests are determined. Then, the prepared data is converted into input data by converting it into the appropriate format to be saved in the database.

**2.2.2 Definition of Requirements**

The process represented by steps 1, 3 and 6 is the requirements definition phase. In the requirements definition phase, legal requirements are defined and, unlike the data processing process, the test methods for these requirements are also selected. In addition to the legal requirements, the analysis and tests to be performed to prove the correctness of the design of the tool and the prototype product, how the results will be analyzed, and the metric values to be used are explained and made into instructions.

**2.2.3 Database**

The process represented by steps 7, 8, 9, 10 and 11 is the preparation of the database. At this stage, the input data is converted into PEGASUS format and prepared for use in analysis and testing. The previously described metric values are matched with the analysis and tests to be applied and collected in a common data pool. In addition, all test scenarios and metric values are compiled, and a test concept is developed.

**2.2.4 Evaluation of Highly Autonomous Driving Functionality**

Steps 12 to 19 are the assessment phase of the high autonomous driving function. In this phase, test concepts and transition criteria are combined and transformed into a logical test case, implementation methods of test concepts are determined, test cases are executed, the data obtained are collected in the data pool and evaluated with predetermined metrics and test results are organized. Risk assessments are also performed.

**2.2.5 Argumentation**

Step 20 is the final processing step of the PEGASUS method and combines the outputs compiled in this step by applying safety argumentation together with the safety statement and risk assessment results obtained. The PEGASUS safety argumentation should be understood as a conceptual framework supporting the safety and validation of higher levels of automation, structured in five layers through structure, formalization, consistency, completeness and conformity.

**2.3 STPA Method**

The Systems-Theoretic Process Analysis (STPA) method was first developed in 2010 by academics from MIT and JAXA/JAMSS universities for theoretical accident analysis of systems consisting of software and hardware components in order to analyze the safety levels of spacecraft at the design stage, determine appropriate measures and reflect them in the design [34]. This analysis method was adapted to the automotive sector in 2013 to determine the safety levels of autonomous vehicles [35].

There are two different applications of the STPA method for the safety of autonomous vehicles. One of them is the ISO/PAS 21448 "Safety of Intended Functionality" (SOTIF) standard. SOTIF was developed by the automobile industry as a safety standard for driving assistance systems. This standard is primarily concerned with reducing to acceptable levels the risks arising from incomplete definition of the function being addressed or from the vehicle being exposed to unexpected conditions that interfere with the operation of sensors and algorithms. SOTIF has been developed as a standard for the safety of driving assistance systems for SAE 0, 1 and 2 level vehicles. The use of the STPA method with SOTIF is possible for SAE 3 and 4 level vehicles. However, it is recognized that SOTIF is a continuation of ISO 26262 and was developed to fill the gaps in the scope of the ISO 26262 standard for driving assistance functions that were not foreseen [36].

The other application area of the STPA method in the safety of autonomous vehicles is the ISO 26262 standard, as addressed in the EU report, which is the reference for the selection of the methods analyzed in this study [19]. The safety of today's conventional road vehicles is verified by the methods described in the ISO 26262 standard. With this standard, different methods such as FTA (Failure Tree Analysis), FMA (Failure Mode Analysis), HAZOP (Hazard and Operability) are used to detect faults that may cause hazards in terms of functional safety of the vehicle. However, these methods are not sufficient for the detection of hazards caused by software errors, human error or component failure.

The STPA method is already in use for analyzing systems consisting of software and hardware components, from spacecraft to naval and air vehicles. STPA promises to overcome the limitations of traditional hazard analysis methods. Since STPA can be used to complement the missing aspects of ISO 26262 in autonomous vehicle assessment, an integrated method has been proposed [37]. Thus, the gaps related to autonomy functions that are not considered in ISO 26262 standard are also completed.

Table 1. Comparison of STPA and ISO 26262 Terminologies [37].

| Term | STPA | ISO 26262 |
|---|---|---|
| Hazard | A system state or set of conditions that, together with a particular set of worst-case environmental conditions, will lead to an accident (loss) | Potential source of harm caused by malfunctioning behavior of the item |
| Malfunctioning behavior | No explicit definition | Failure or unintended behavior of an item with respect to its design intent |
| Failure | No explicit definition Note: A failure in engineering can be defined as the non-performance or inability of a component (or a system) to perform its intended function | Termination of the ability of an element, to perform a function as required Note: Incorrect specification is a source of failure |
| Accident | An undesired or unplanned event that results in a loss, including loss of human life or human injury, property damage, environmental pollution, mission loss, etc. | No explicit definition |
| Harm | No explicit definition | Physical injury or damage to the health of persons |
| Hazardous event | No explicit definition | Combination of a hazard and an operational situation |

ISO 26262 describes methods where the hazard analysis of the vehicle is examined separately for subsystems or risks. On the other hand, although the driver plays a critical role in the driving of the vehicle, there is no clear understanding of whether the driver should be considered as a component of the vehicle in terms of functional integrity [37]. In STPA, on the other hand, all tasks undertaken by the driver are attributed to the vehicle and are therefore clearly included in the evaluation domain of the STPA method.

As the STPA method brings changes in driver-specific understanding, the terminologies adopted in ISO 26262 also need to be modified. Table 2.1 compares the terms that differ. By explaining why the terms in Table 2.1 differ between STPA and ISO 26262, it will be clearer how STPA complements which aspects of ISO 26262. First of all, while the term hazard is evaluated in ISO 26262 in terms of the subsystems of the vehicle, in STPA, it is not limited only to the subsystems of the vehicle, but is addressed in a much broader perspective in terms of the environmental conditions in which the vehicle is in driving conditions, other actors in traffic and the instantaneous changes of all these environmental conditions or the behavior of the actors in traffic.

Although the term malfunctioning behavior is defined in ISO 26262 as the loss of functionality or improper operation of any component of the vehicle that could cause an accident, there is no clear definition of this term in the STPA method. The same can be said for the definition of malfunction. ISO 26262 defines failure as the cessation of a component's ability to fulfill a function that it is required to fulfill, but there is no clear equivalent for this term in the STPA method [37]. The lack of a clear definition of these two terms in the STPA method is due to the fact that, as in the case of the term hazard, the evaluation area is expanded beyond the components of the vehicle to include all environmental elements in which the vehicle plays a role. According to ISO 26262, the cause of an accident is considered to be the failure of one of the components of the vehicle, whereas STPA considers the cause of an accident to be anything that may affect the safe operation of the vehicle, including environmental influences, so the concept of failure as the cause of a potential accident loses its meaning.

When examining the definition of an accident, when we consider that the focus of ISO 26262 is on hazards based on the possibility of failure of the components that make up the vehicle, the realization of the accident event falls outside of ISO 26262. Since the focus of STPA includes the realization of the hazard outside the vehicle, an accident is also covered. On the other hand, the definition of accident in STPA is similar to the concept of harm in ISO 26262, but based on a much broader scope. Whereas ISO 26262 deals with damage to people, STPA is not limited to people but includes any loss, including damage to persons or property, environmental pollution, and inability to fulfill the intended task of transportation. The concept of hazardous incidents is defined in ISO 26262 as situations that will result in damage, while the definition of accident in STPA refers to events that will result in damage. There is no clear definition of the concept of hazardous incident in STPA [37].

A comparison between STPA and ISO 26262 is also made in terms of safety coverage [38]. The safety coverage of ISO 26262 is based on the HARA method. A representation comparing the scopes of these two methods is given in Figure 13.

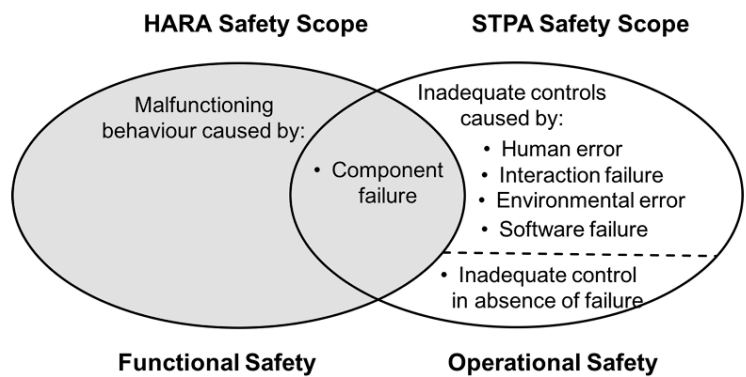

**Figure 13.** Comparison of STPA and ISO 26262 Scopes [38]

Comparing the scope of safety between HARA and STPA, HARA considers the source of danger to be a malfunctioning situation resulting from an equipment failure, while STPA considers it to be any situation that may lead to loss of control, such as human error, interaction error between the elements involved in traffic, unforeseen environmental conditions or software failure.

**2.3.1 Steps of STPA Method**

According to all these explanations, STPA is considered as a complementary method that extends the scope of ISO 26262 to address the additional issues brought by the autonomy functions of the autonomous vehicle. Thus, STPA is envisaged to be used within the scope of ISO 26262 and the steps to be followed for this use are explained as follows [38].

1. Apply STPA Step 0 (Fundamentals Analysis):
   1.1. Identify Accidents and system-level hazards.

1.2. Identify the high-level system safety constraints.
1.3. Draw the control structure diagram of the system.
2. Use the results of STPA Step 0 to define an item and item information needed (e.g. purpose, content of item, functional requirements etc.). The control structure diagram in STPA Step 0 shows the main components which form a system under analysis. This diagram contains information to help the functional safety engineer to define an item and its boundaries.
3. Use the list of hazards, accident, the high-level system safety constraints identified in STPA Step 0 as an input to the HARA approach.
4. Apply the HARA approach:
4.1. Determine the operational situations and operating modes in which an item's malfunctioning behaviour may lead to potential hazards.
4.2. Classify the hazards identified in Step 0 based on the estimation of three factors: Severity (S), Probability of Exposure (E) and Controllability (C)
4.2.1. Identity the hazardous events by considering the hazards in different situations.
4.3. Determine ASIL (Automotive Safety Integrity Levels) for each hazardous event by using four ASILs: A (the lowest safety integrity level) to D (the highest one). If the hazardous event is not unreasonable, we refer to it as QM (Quality Management).
4.4. Formulate the safety goal for each hazardous event.
5. Use the hazardous events, safety goals, situations and modes as input to the STPA Step 1.
6. Apply STPA Step 1 to identify the unsafe control actions of an item.
7. Apply STPA Step 2 to identify the causal factors and unsafe scenarios of each unsafe control action identified in STPA Step 1.
8. Use the results of STPA Step 1 & 2 to develop the system functional safety concept and safety requirements at this level.

The application process of the STPA method is shown as a flow diagram in Figure 14. According to Hu, STPA analysis starts with specific accidents, analyzing hazardous situations and safety constraints required to eliminate the negativity that will cause hazards in the safety related behaviors of the system [39]. Then, unsafe behaviors in the control structure of the system should be identified and the necessary safety constraint in the control of the vehicle is obtained by analyzing the causes of these behaviors.

## 2.3.2 Using STPA with ISO 26262

We have already explained that the STPA method was developed to be used in combination with the ISO 26262 standard for the safety of vehicles. The distribution of the steps of STPA between ISO and STPA is represented in Figure 14 [38].

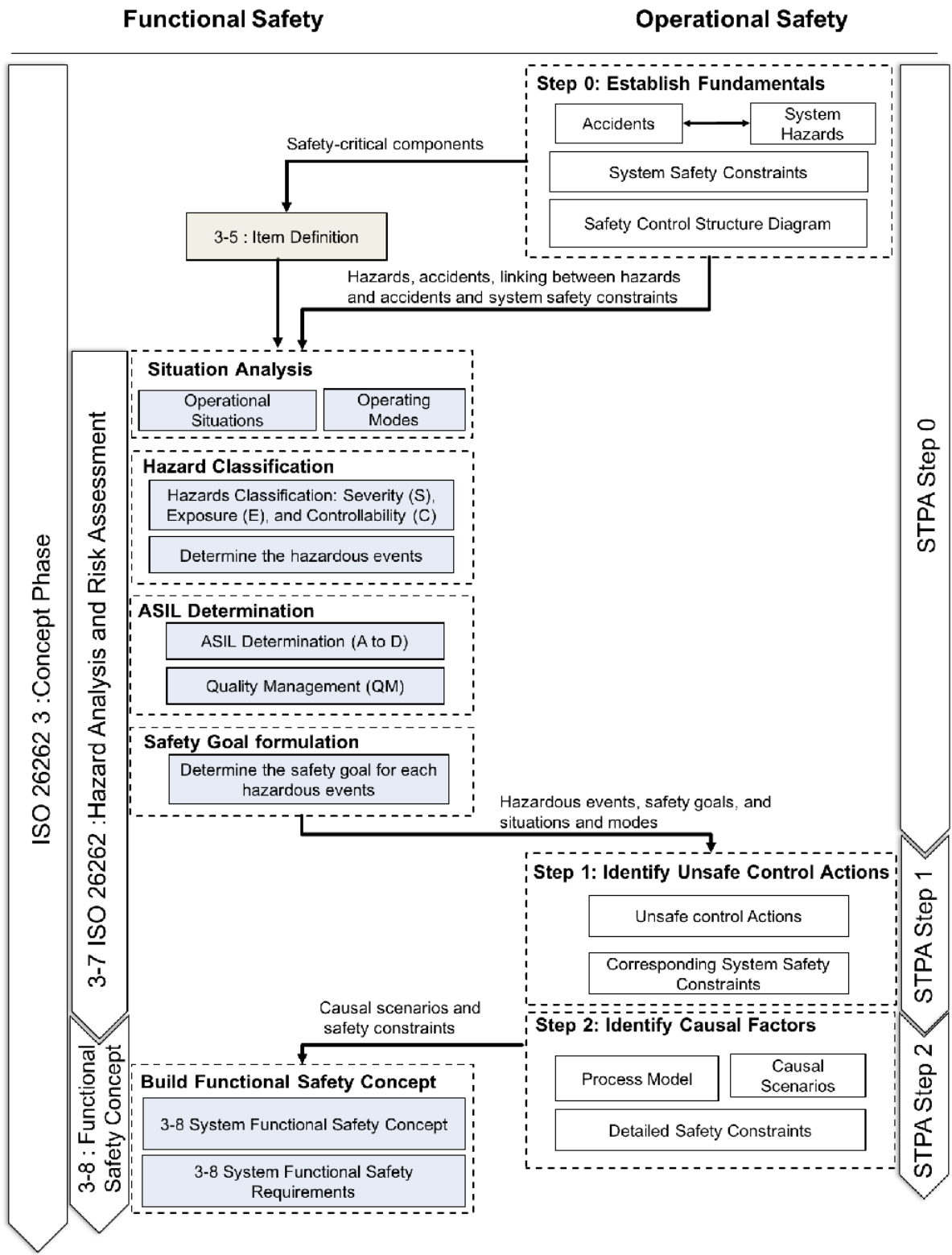

**Figure 14.** Integration of STPA into ISO 26262 [38].

The HARA (Hazard Analysis and Risk Assessment) process in ISO 26262 consists of four steps. In these steps, hazards are analyzed, hazards are identified, classified and safety objectives are determined. However, the risks analyzed in these steps address the hazards that may arise from the subsystems of the vehicle. However, in an autonomous vehicle, in addition to the vehicle's subsystems, driving methods and preferences, driving behaviors of the vehicle's autonomy function, climate and environmental conditions, as well as hazards that may arise from risk sources such as the behaviors of other vehicles playing a role in the driving environment should also be addressed. STPA fulfills this need and ensures that these risks are addressed in Steps 1 and 2 in addition to HARA.

## 3. MODELING OF THE CERTIFICATION PROCESS AND A COMPARISON OF PROMINENT METHODS

In this section, the certification process of an autonomous vehicle is modeled hierarchically, the stages at which RSS, PEGASUS and STPA methods can be used in this process are explained, and the relationship between the methods is explained within a taxonomy comparing the methods in terms of application areas, possible user stakeholders and applicability.

Figure 15 schematically illustrates the certification, type approval and licensing stages of autonomous vehicles. In the figure, the design, production, testing, certification and type approval process [40] of railway vehicles are adapted for autonomous vehicles. The first phase of Figure 15, the requirements definition phase, consists of gathering requirements from two sources. The first one is the requirements set by the vehicle manufacturer, taking into account the expectations of the market and the economic acceptability to consumers if these expectations are realized. These requirements may consist of basic design features of the vehicle such as size or maximum speed, or they may consist of comfort functions such as air conditioning, type of steering control mechanism, seat heating, interior lighting, or the quality of the materials to be used. The second source is national and international legislation. Following the requirements determined in the first stage, the requirements of the national and international rules determined for the type of vehicle should be collected and all the requirements that the vehicle must meet should be determined.

The RSS method can answer some of the questions about the autonomy functions that the vehicle should fulfill through the scenarios it is expected to overcome. In the PEGASUS method, the certification process is clarified with a systematic model and considering that the process starts with the requirements definition phase, this method also provides responses to the requirements definition phase.

In the 2nd stage, concept designs should be made based on the requirements determined. These designs constitute the designs that will be taken as reference in all detailed designs of the vehicle, from the structural design of the vehicle to the system architectures related to each subsystem and the determination of the interfaces of the subsystems and the formation of system integrity designs.

In the 3rd stage, the hardware and materials suitable for the requirements determined by the concept designs are determined, and the structural designs and system architectures according to the concept designs are updated by detailing them with the determined hardware. At this stage, the communication architectures are also updated, taking into account the characteristics of the hardware specified for use in the system architectures, the area of use and the distances of the hardware. While preparing the hardware infrastructure, the appropriate hardware required for the realization of the autonomy functions that need to be fulfilled in the scenarios considered in the RSS and PEGASUS methods should be determined.

In the 4th stage, the software infrastructure is prepared by determining the software, programming languages and algorithms that will be used to ensure the functions that each subsystem must fulfill in the system architectures whose hardware and communication details are determined and the integration of these subsystems in a compatible and synchronous manner. While preparing the software infrastructure, the algorithms required for the realization of the autonomy functions that need to be fulfilled in the scenarios considered in the RSS and PEGASUS methods are prepared.

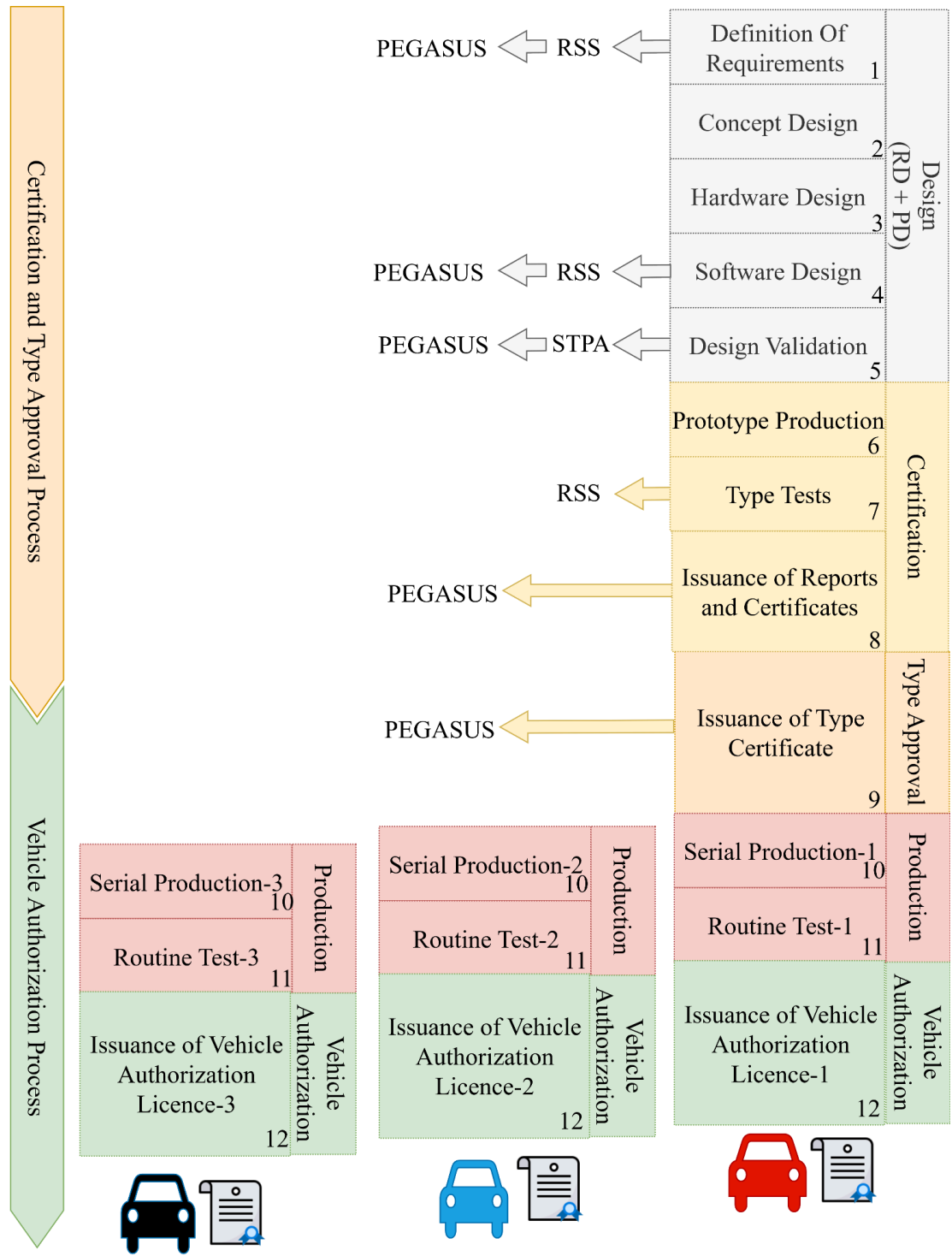

**Figure 15.** Roles of RSS, STPA and PEGASUS Methods in Certification Process.

In the 5th stage, it is verified that the designs, which are completed with structural, hardware, communication and software details, can fulfill the specified functions in accordance with the targeted performance criteria with a trouble-free integration of each subsystem with each other. This verification is carried out using simulations and analysis methods specified in international standards for each function. After confirming that the simulation and analysis results provide the expected results, prototype production begins. The STPA method is used for hazard analyses related to autonomy functions that are not covered in ISO 26262 in hazard analyses, and the PEGASUS method is used to ensure the traceability of the verification by all parties, including official authorities, within the certification process.

In the 6th stage, a prototype of the hardware, subsystems and integrated system to be used in line with the concept, hardware, communication and software designs is produced.

At stage 7, field and laboratory tests and examinations of the equipment, subsystems and integrated system produced are carried out. It should be taken into consideration that the tests carried out should be carried out in accredited laboratories, field tests and examinations should be carried out under the supervision of conformity assessment bodies authorized according to their nature, and the analysis of the test results should be confirmed by the conformity assessment bodies.

In the 8th stage, the type test results are evaluated by the conformity assessment bodies according to the nature of each test or examination, and it is certified that the requirements specified in national and international legislation are met.

At stage 9, the certificates specified in national and international legislation are submitted to the official authorities with the type approval application for placing on the market. If it is found appropriate by the official authorities with the examinations to be made on the certificates and reports submitted, it is allowed to start serial production with the type approval to be issued regarding the design of the vehicle.

After this stage, stages 10 and 11 are repeated for each mass-produced vehicle. At stage 10, the vehicle is produced in accordance with the type approval in production facilities certified to comply with the technical and administrative requirements set out in national and international legislation. In the 11th stage, the conformity of the vehicle produced with the type approval requirements is verified by routine tests. In the 12th stage, the vehicle license is issued by applying to the relevant official authority for licensing together with the evidence that the vehicle produced complies with the type approval requirements. The vehicle authorization required for the vehicle to be put into use can be considered as the licensing of the vehicle.

## 3.1 Application Areas of the Methods

As a result of the research conducted within the scope of this study, it was understood that the RSS method was developed to determine the basic principles related to the autonomy functions of the vehicle. In this respect, it should be taken as a basis in product development (P&D) studies. On the other hand, whether the produced vehicle meets the principles determined by the RSS method should be tested with scenarios to be produced in line with these principles. This test mechanism is explained in detail in the PEGASUS method. The PEGASUS method explains all the process steps one by one and gathers the entire process from the concept design stage to testing, certification and type approval by the official authorities into a mechanism. The STPA method describes the steps required to determine the risks that may arise from the vehicle or external environmental conditions or actors in traffic due to the autonomy functions of the vehicle and the measures to be taken to reduce these risks to an acceptable level. By this way, It responses to the gaps that need to be filled in the ISO 26262 standard, which is currently used for conventional cars.

## 3.2 Possible User Stakeholders of the Methods

Since the RSS method determines the principles related to the autonomy functions of the vehicle, it will be used by the P&D departments of vehicle manufacturers. On the other hand, since these principles will also be used in determining the test scenarios of the vehicle, the official authorities in charge of determining the requirements for type approval and preparing regulations and the test centers that will test compliance with these requirements will also be among the possible users of the RSS method.

In the PEGASUS method, the design, production, testing and certification processes of the autonomous vehicle are addressed as a whole, so subsystem providers, vehicle manufacturers, test centers and authorities are the potential stakeholders that will use this method.

Risk assessment bodies are potential users since the STPA method completes the identification of the risks posed by autonomy functions, which are missing in the ISO 26262 standard where the conventional vehicle is tested, and the determination of the necessary measures to reduce the risks to an acceptable level.

**3.3 Applicability of the Methods**

When assessing whether the development of all three methods is sufficiently complete for use in the certification process of an autonomous vehicle, it is evident that the RSS method stands out as adequately developed for application. This is due to its role as a reference in most scientific and industry studies on autonomous vehicles to date. Consequently, the RSS method is currently considered applicable.

The PEGASUS method is a project involving automobile and subsystem manufacturers and researchers from the scientific community. Although the aims, objectives and concepts of the project have been explained, gaps in the technical regulations still need to be addressed for them to be put into practice. On the other hand, to complete the PEGASUS project, the necessary regulations must be finalized and put into practice. It is known that France and Germany are actively working on this issue. For these reasons, regulatory efforts to ensure the method's applicability are ongoing.

The STPA method was developed to adapt ISO 26262 for the use of autonomous vehicles, and academic studies have particularly focused on the assessment of subsystems. Given that the method is clear and has practical applications, it is considered applicable.

**3.4 The Role of the Methods in Certification Process**

Although RSS, PEGASUS and STPA have been published as methods being developed by the Council of Europe for the certification of autonomous vehicles, it has been seen that these three methods are not alternative methods to each other, but methods that respond to the needs in three different areas. RSS clarifies the determination of design and test criteria, PEGASUS clarifies the process from design to type approval and the roles of the actors in the process, and STPA clarifies the hazards analysis of autonomous functions. Thus, they are complimentary methods.

**4. CONCLUSION**

Today, even though autonomous vehicles have been commercialized and offered to users based on local legislation in some countries, studies on the national and international regulatory requirements of autonomous vehicles have not yet been finalized to clarify the safety requirements, and which methods will be used to test according to these requirements.

In this study, the RSS, PEGASUS and STPA methods discussed in the "New Approaches for the Certification of Autonomous Vehicles" report published by the Council of Europe in 2020 were examined, and the certification process of autonomous vehicles was turned into a model by explaining at which stages, in which areas and by which actors these methods should be used in the certification and type approval process of autonomous vehicles.

The most important deficiency observed in the methods that are prominent in the certification process of autonomous vehicles is that the methods examined in this study do not foresee that changes of unexpected conditions of the road as a result of climatic conditions, landslides, accidents or maintenance and repair works or may not fulfill the autonomy functions of the vehicle at the predicted performance conditions in the scenario-based tests. In such a case, it is evaluated that the performance conditions and detection capabilities for unexpected conditions of the autonomous vehicle should be improved.

Similarly, it should be taken into consideration that the factory performance data based on the formulas used in the design of the vehicle's functions may change over time as a result of the wear and tear of the vehicle's tires or consumables that will affect the performance conditions of the braking performance, and it should be taken into account that after the vehicle prototype successfully completes the type test phase, it will not be able to provide the performance obtained during the type test phase in the scenarios that the serial produced vehicles are expected to face during their lifetime. In order to overcome this problem, it is necessary to continuously compare the predicted performance values with the realizations in the field and to reflect the current performance values of the vehicle in the formulas used in the design. Therefore, it is considered that adhesion should be included in the mathematical modeling in the formulas described in the first and second rules of RSS, anekid the artificial intelligence based managing the vehicle should recalculate and update the maximum braking distance according to the adhesion coefficient in the field by measuring the actual realization of the braking command to the braking subsystem.

In future studies on the design, certification and type approval of autonomy functions of autonomous vehicles, it is necessary to complete the deficiencies of the methods examined in this study, as explained above, and to continue to work on the preparation of national and international regulations by making the certification and type approval process an official model.

**ACKNOWLEDGEMENT**

The study was partially supported by ROYAL CB SBB 2019K12-149250.